\documentclass[]{spie}  

\usepackage{amsmath,amsfonts,amssymb}
\usepackage{graphicx}
\usepackage{svg}
\usepackage[colorlinks=true, allcolors=blue]{hyperref}
\usepackage{textcmds} 
\usepackage{enumitem}
\usepackage{comment}

\title{Improving generalization with synthetic training data for deep learning based quality inspection}

\author[1]{Antoine Cordier}
\author[1]{Pierre Gutierrez}
\author[1]{Victoire Plessis}

\affil[1]{Scortex, 22 rue Berbier du Mets, Paris, France}

\begin{document} 
\maketitle


\begin{abstract}
Automating quality inspection with computer vision techniques is often a very data-demanding task. Specifically, supervised deep learning requires a large amount of annotated images for training. In practice, collecting and annotating such data is not only costly and laborious, but also inefficient, given the fact that only a few instances may be available for certain defect classes. If working with video frames can increase the number of these instances, it has a major disadvantage: the resulting images will be highly correlated with one another. As a consequence, models trained under such constraints are expected to be very sensitive to input distribution changes, which may be caused in practice by changes in the acquisition system (cameras, lights), in the parts or in the defects aspect. In this work, we demonstrate the use of randomly generated synthetic training images can help tackle domain instability issues, making the trained models more robust to contextual changes. We detail both our synthetic data generation pipeline and our deep learning methodology for answering these questions.
\end{abstract}

\section{Introduction}
\label{sec:intro}  

The lack of availability of training data is one of the biggest challenges when applying deep supervised learning to quality inspection. Annotating defective data not only is a costly and difficult task, but also a highly inefficient one: gathering enough defective data with high variability may in some cases simply not be possible, given the rarity of some defects. This is especially true when applying computer vision techniques to video frames, which are highly correlated with one another. This issue contrasts with more standard deep learning vision tasks such as ImageNet\cite{5206848}, where each class is represented with numerous images of great diversity in the training set. Consequently, the learnt models are often very sensitive to distribution changes, such as lightning condition or view angle changes. Data augmentation may be used to improve the models robustness to such changes, but has its own limitations\cite{azulay2018deep}.

To overcome the data availability problem, an encouraging approach is to automatically generate labeled training data using a synthetic data generation pipeline\cite{tremblay2018training}. Indeed, generating a wide distribution of randomized scenes with defective parts may not only help the model learn the defects on the original real domain better, but should also improve their generalization capability to unseen changes of distributions: this paradigm is commonly referred to as domain randomization\cite{tobin2017domain}. As a result, deep learning networks may be trained more effectively, leading to more robust and adaptive models. This idea has been effectively used in the autonomous driving industry\cite{Pouyanfar_2019_CVPR_Workshops, prakash2019structured}.

Following our work on assessing the usefulness of synthetically generated training data for deep learning based quality inspection\cite{DBLP:journals/corr/abs-2104-02980}, we here aim to evaluate the impact of the use of synthetically generated data on the generalization capability of deep convolutional neural networks (CNNs) trained on the same kind of defect detection task. In this paper, we define four changes of distributions for our data, and evaluate models trained with or without simulated data on test sets which reproduce these chosen changes of distributions. These test sets will be referred to as \textit{transfer test sets}, and include an acquisition system (cameras and lights) change, two part aspect changes, and a defect aspect change. 
For this study, the parts to be inspected are car doors from one of our clients, while the only targeted defects are breaks.

We first review works from the literature that are related to ours in section \ref{sec:related_work}. We then detail our methodology in section \ref{sec:methods}, for both the simulation data generation pipeline and the training of our deep learning networks. Our results are presented in section \ref{sec:experiments}. We demonstrate that while simulated data may not improve performances on the original real domain significantly, its role in making the model more robust to unseen distribution changes is substantial, as shown by the improvement of performance observed on the transfer test sets. During this work, we experimented with various domain adaptation techniques (namely DANN\cite{ganin2016domain} and ASS\cite{wang2020alleviating}) in order to help bridge the gap between the real and simulated image domains, but without success. Consequently, we will not be presenting the results related to our domain adaptation experiments in this paper.

\section{Related work}
\label{sec:related_work}  

Automated visual inspection via the use of \textbf{deep learning} often relies on supervised learning\cite{dong2019pga}. In order to classify and localize defects on visual data, convolutional neural networks (CNNs) are trained on images, most of the time using either semantic segmentation networks (such as U-Net \cite{ronneberger2015u}), or object detection architectures (such as YOLO\cite{redmon2016you} or RetinaNet \cite{lin2017focal}). However, one of the drawbacks of deep supervised learning is that it requires large quantities of data for training. Unfortunately for the quality control usecase, obtaining images of defective parts can be difficult, since defectiveness may be a rare event. In other words, collecting enough data to properly learn to detect defects may not always be possible. To tackle this issue, one can move from supervised to unsupervised anomaly detection techniques \cite{bergmann2019mvtec}, or make use of simulated data to enrich the training set \cite{zambal2019end}.

For our literature review on the the use of \textbf{simulated data for deep learning}, we repeat some elements from our previous work on the topic.\cite{DBLP:journals/corr/abs-2104-02980} The reader may refer to this paper for a more comprehensive review. Simulated training data use has been widely studied for object detection in images\cite{tremblay2018training}, most specifically in the fields of robotics \cite{tobin2017domain}, as well as for autonomous driving \cite{li2019aads}. They are particularly useful for handling tasks where human labeling is hard or costly. This includes pose or depth estimation \cite{chen2016synthesizing,sundermeyer2018learning,langlois2019fly}, and deep learning-based rendering \cite{deschaintre2018single}. Simulated data does not only lower the cost of human annotation, but also allows to learn theoretical modes for which no real data exist. Nevertheless, it is common for models trained on simulated data to fail at generalizing well on real data. This phenomena is known as the \textit{domain gap} problem. 

Three techniques help solving this issue: improving photorealism (bringing the synthetic domain closer to the real one), synthetic domain randomization (randomizing the synthetic domain enough so that it covers the real one), and domain adaptation (train the model in a way that the two domains align with one another). \textbf{Improved photorealism} is essential for learning strong detectors \cite{movshovitz2016useful}. It has recently been shown that rendering 3D models in physically realistic positions can have a significant impact on performances\cite{hodavn2019photorealistic} compared to randomly rendering the same models in random positions on random backgrounds\cite{dwibedi2017cut}. However, in order to generate simulated data diverse enough for learning, photorealism has to be coupled with domain randomization \cite{tobin2017domain,prakash2019structured}.
The idea behind \textbf{domain randomization} \cite{tobin2017domain} was initially to randomize the simulation process so that the real domain could become a subset of the simulated one. This has successfully been applied to autonomous driving \cite{tremblay2018training} and industrial applications \cite{lee2019automatic}. The concept has now been extended to “automated domain adaptation” \cite{akkaya2019solving} (in which the randomization increases over time in a curriculum learning fashion) and to “structured domain randomization” \cite{prakash2019structured} (for which domain randomization does not come at the cost of photorealism). To bridge the domain gap between synthetic and real data domains, \textbf{domain adaptation} techniques can finally be used. We refer the reader to the review by Toldo, Marco, et al. \cite{toldo2020unsupervised}. In our experiments, we tried DANN \cite{ganin2016domain} and ASS \cite{wang2020alleviating} methods, but do not exhibit detailed conclusions, since no significant results were achieved on this subject.

In this article, we focus on combining both real and simulated data at the same time, a topic already covered in some articles\cite{tremblay2018training,wang2020alleviating}. More particularly, we are interested in generalization of models trained with simulated data to new image distributions, e.g. changes of acquisition systems or part references. This topic has strong ties to generalization under covariate shifts, as well as to neural networks  \textbf{robustness against natural perturbations}. Several datasets have been proposed to evaluate networks generalization and transferability to other domains. For example, ImageNet-A and ImageNet-O \cite{hendrycks2021natural} are composed of “natural adversarial examples”. On the contrary, ImageNet-C and Imagenet-P \cite{hendrycks2019benchmarking} were later created synthetically from ImageNet, via the use of common pertubations like noise, motion blur, random contrast, and elastic deformations. Finally, Imagenet-R \cite{hendrycks2020many} is composed of images of ImageNet objects with different renditions (paintings, cartoons, sculptures, origamis).

Several solutions have been proposed to tackle robustness to natural perturbations. First, networks architectural modifications. Larger and deeper networks have been proven to be more robust \cite{hendrycks2019benchmarking}. Leveraging attention, through the use of SE and SK blocks \cite{hendrycks2021natural, lee2020compounding} has also been proven to be useful in some cases. Dedicated architectural layers such as BlurPool \cite{zhang2019making} can be used to tackle the fact that modern neural networks are not translation invariant nor equivariant.
Second, data augmentation based techniques. Common data augmentations such as CLAHE \cite{hendrycks2019benchmarking} or Mixup \cite{lee2020compounding, zhang2017mixup} may help. Stability training \cite{zheng2016improving} inspired techniques such as Augmix \cite{hendrycks2019augmix} or Auxiliary training \cite{zhang2020auxiliary} can also work. However, it has been proven data augmentation has trouble generalizing outside the training data manifold, and thus do not provide general network invariances. \cite{azulay2018deep}. Close to this idea, self-supervised learning with rotations prediction \cite{hendrycks2019usingself} can also improve robustness.
Last but not least, transfer learning \cite{hendrycks2019using, taori2020measuring}. Transfer learning should preferably be applied using deep and large networks trained with very large corpuses of images coming from external domains.\cite{kolesnikov2019big}. 

Studies have shown these techniques are not effective for all cases, and that there is no such thing as a silver bullet for making CNNs more robust to perturbations \cite{hendrycks2020many, taori2020measuring}. Therefore, the robustness of neural networks is still a very open research topic. In this article, we propose improving generalization via the use of synthetic training images. To some extent, this can be seen as an extension of data augmentation techniques.

\section{Materials and methods}
\label{sec:methods}  

\subsection{Synthetic training data generation}
\label{subsec:data_generation}
We designed a synthetic image generation pipeline based on Blender\cite{blender} and its physically-based rendering (PBR) engine Cycles. Our software allows us to randomly generate large batches of realistic images with great variability, and can be controlled with a simple configuration file. While realism is ensured via advanced physically-based modelling and rendering options (e.g. Monte Carlo ray tracing\cite{lafortune_mathematical_nodate}, IES lights\cite{noauthor_illuminating_nodate}, PBR materials\cite{cook_reflectance_nodate,burley_physically_nodate}, ...), variability is achieved through the randomization of several simulation parameters (such as position, orientation, and other characteristics for lights, cameras, parts and defects). Our core idea is to couple photorealism with domain randomization\cite{tobin2017domain,prakash2019structured} to generate realist, but diverse enough images of defective car doors.

\subsubsection{Environment generation}
\begin{figure}
    \centering
    \includegraphics[width=12cm]{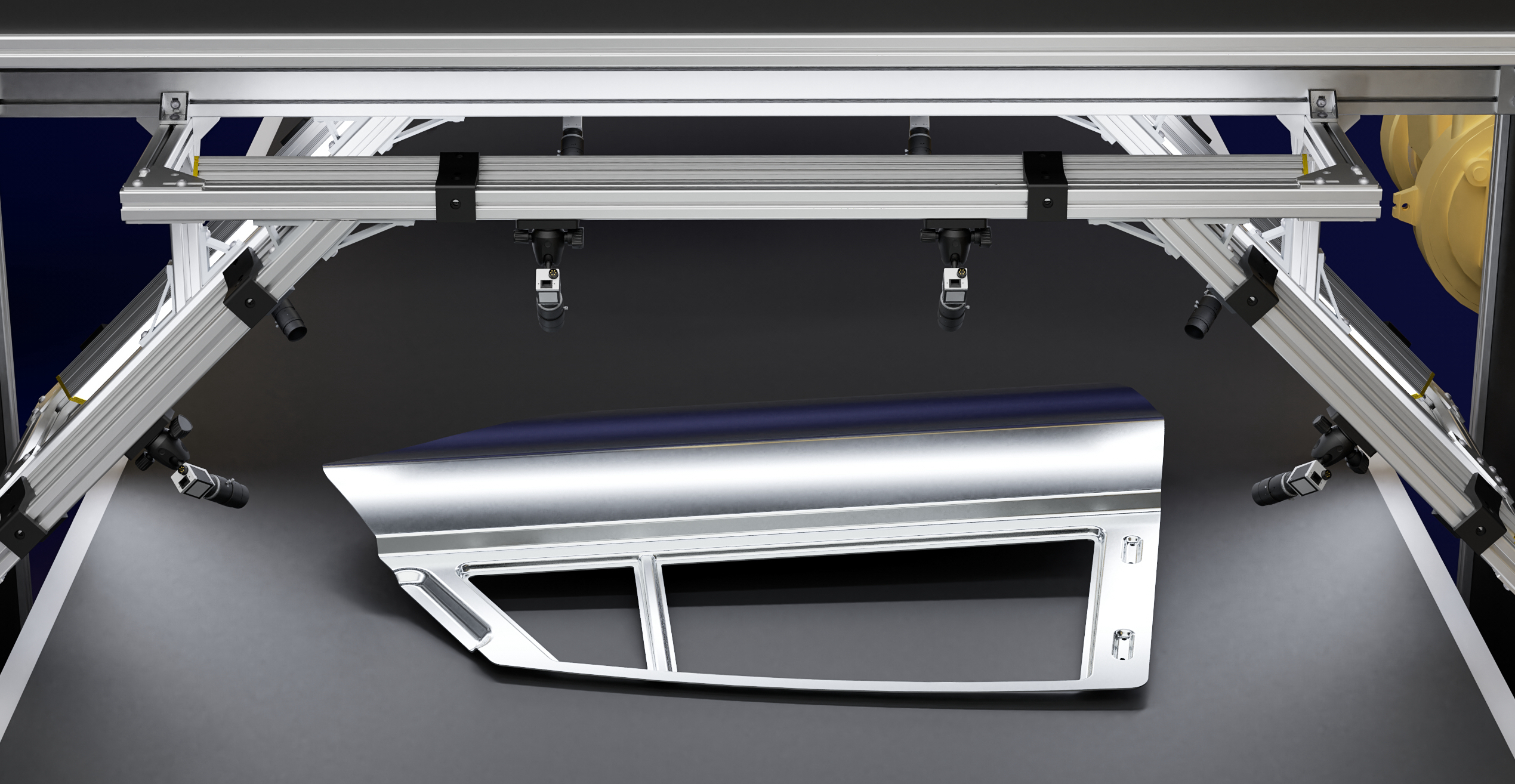}
    \caption{The default simulated environment for our scene, with an additional positioned part.}
    \label{fig:stage_set}
\end{figure}

First, we design a bare environment for our scene, which is a 3D simulated stage based on the real factory setup: a conveyor belt, over which an ark supporting nine cameras is placed. Default positions and orientations are chosen for the opto-electronic elements (cameras and lights) based on the real setup, which are then placed on the stage. An overview of the default environment with an additional positioned part can be seen in Figure \ref{fig:stage_set}.

For both cameras and lights, pose (i.e. orientation and position) is randomized. For cameras, we also randomize the focus blur (via the working distance and aperture). For lights, we additionally randomize light power and color. Environment randomization is then finalized by randomizing the texture of some elements of the stage (e.g. conveyor belt).

\subsubsection{Parts generation}

After importing the CAD model for the parts of interest (which were given by our client), one to several parts are then selected to be positioned on the stage, which is an horizontal plane. Because a set of known physical equilibrium positions has been attributed in advance to each part with regards to this plane, it is possible to randomly pick one of these equilibrium positions for each selected part, before placing it on the plane. The part is placed on the plane with an additional random translation and rotation in the plane to ensure pose randomization. To make sure no unrealistic overlaps between the parts and the rest of the environment can happen, multiple positioning attempts are made until a satisfying result is achieved. The part material and textures are randomized using procedural texturing.

\subsubsection{Defects generation}
\label{defects_generation}

A set of defect texture maps (normal, height, roughness, and occlusion maps) is generated in advance with a procedural texture generator. Once generated, these texture maps can then be combined in Blender for randomly placing defects on the parts. We define our simulated defects as a combination of up to three types of textures: deformation, encrustation, and hole textures. The exact combination and characteristics of these textures will depend on the desired defect.

\begin{figure}
    \centering
    \includegraphics[width=17cm]{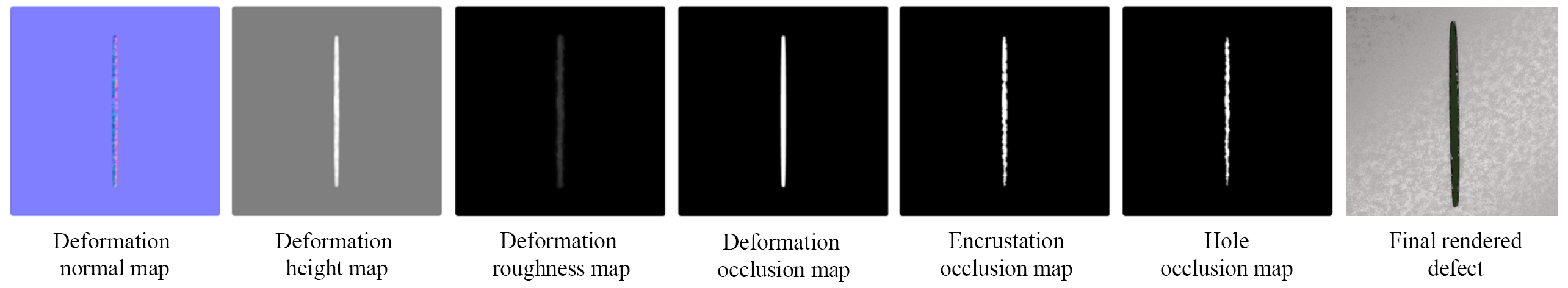}
    \caption{The different texture maps used for simulating a break, with the final rendering on the far right.}
    \label{fig:cassure_text}
\end{figure}

\begin{itemize}
\item Deformation textures are used for defects which are defined as a physical deformation of the part, such as shocks or scratches. In these cases, we create four texture maps (normal, height, roughness maps, as well as an occlusion map to select the deformed zone) to apply.
\item Encrustation textures are used for defect which are defined as a distinct material change or insertion on the part, such as craters, burns, or marks. In these cases, a unique texture mask map is created, and a new PBR material is applied locally using the mask.
\item Hole textures are used for defects which are defined as a zone where the original material from the part is missing, such as holes or cracks. In these cases, a texture mask map is used for removing the material where desired.
\end{itemize}

In our case, the defect of interest is a break. To simulate it, we combine all three approaches mentioned above, as shown in Figure \ref{fig:cassure_text}. When placing a defect on a part, both its position on the part mesh and size are randomized. When randomizing the size of the defect, texture maps of the latter are scaled accordingly.

\subsubsection{Rendering scheme}
Once parts and defects are placed, a content verification algorithm is ran for each camera field of view to decide whether the generated scene should be rendered for the given camera: this allows to only render useful images, since rendering time represents the main compute cost of our pipeline. The algorithm consists in verifying the presence of at least one visible part with at least one visible defect for the given camera. Visibility of a part is decided following geometrical considerations: at least one vertex of the part model should fall within the field of view of the considered camera. Visibility of a defect is decided following a specific rendering of the scene, where only the selected defect reflects light: this is possible thanks to the available texture mask attached to each defect, as explained in \ref{defects_generation}. The result will be an image where only the defect is visible, while the rest is dark. After binarization of this image, a visibility mask is created for the defect, and a threshold can then be applied to the size of this mask to decide whether the defect is considered visible or not. Note that this visibility mask will also serve for generating the label annotation associated to this defect. To avoid mistakes, if a defect is considered not visible, it will simply be removed from the part before rendering. Finally, an image will be rendered only if at least one visible defect is present on the part.

Thereby, we were able to generate large batches of synthetic images of defective parts. Ensuring that the generated images follow a similar distribution to that of the real ones may be one of the hardest challenges of the described approach. In practice, iterating on the simulation parameters can become a long and tedious process. Additionally, one needs to define adequate metrics for measuring the proximity between simulated and real images. While experimenting, we iterated on the size of the synthetic defects, after understanding that the size of the defects we were initially generating was too big. Such an iteration had a significant impact on the performance of the simulated models (data not shown). However, it is currently not possible to confidently predict the impact a given iteration will have on the performances of the future model before training the latter. This makes iterating laborious, even though it probably is a crucial element of the overall approach.

\subsection{Defect detection with deep learning}
\label{subsec:defect_detection}

In order to detect defects, we rely on supervised deep learning. The detection task is framed as a binary segmentation problem for training, but is evaluated using aggregated boxes for computing recall, precision and mAP, since it is closer to the manufacturer metrics of interest. Because the parts are rather large (2 meters long) and the defects may be relatively small (a few centimeters), we use input images of high resolution (2048 x 2592). No resizing or cropping is used as inference pre-processing. Because of this, we rely on custom architectures in order to ensure real-time inference with a high frame rate. These architectures are inspired from the literature (e.g. VGG, ResNet, U-Net, or RetinaNet\cite{lin2017focal}) but have thinner and shallower backbones.

In order to improve performance and generalization, we apply data augmentation to all training images, including the synthetic ones. We use random noise, blurring, contrast, color jittering, cutouts, flips, zooms, translations, rotations, and shears.

\section{Experiments}
\label{sec:experiments}


Similarly than in our previous work on synthetic data generation\cite{DBLP:journals/corr/abs-2104-02980}, we will answer the following questions:
\begin{enumerate}[label=(\alph*)]
\item Can we achieve good performances on real data with simulated images alone? 
\item Can we improve (a) by using additional healthy real images? This is an important question since healthy parts (hence images) are easily collectable in large quantities, as opposed to defective parts, which we may not have at our disposal in case the defect is a rare event.
\item Can we leverage simulations as additional inputs to improve over models trained with real data only? 
\end{enumerate}

In all of these experiments, we assess whether simulations can improve generalization capability over unseen distributions, e.g. robustness against part and defect variations, or against external condition changes. This is possible because of the applied domain randomization when generating the simulated data.

\subsection{Training and evaluation data}
We generate 9413 simulated images, which are split into a training (8871 images) and a test (542 images) set. The CAD model used to generate the simulated parts corresponds to the same part reference that is used for the real training images. Each generated image contains at least one defect. It is worth noting that the generation of this data only took three days, which is notably faster than what it would have taken for a single person to annotate an equivalent amount of defective real images (which may not necessarily be available). Note also that we fine-tune our simulation parameters so that we generate defects that have comparable size with real ones.

We also collect real images from our client parts directly from the production line, in the factory: we collect 10000 images for the real training set, each of which also contains at least one defect. This allows us to have comparable simulated and real training sets in terms of number of images and defects. Additionally, 10000 extra real healthy (i.e. without defects) images are collected, all coming from the same distribution than the defective real images (same acquisition system, same part reference, but healthy). Note that the amount of extra healthy images corresponds to the number of images present in the real training set to ensure comparability of trainings (see next section).

\begin{figure}
    \centering
    \includegraphics[width=13cm]{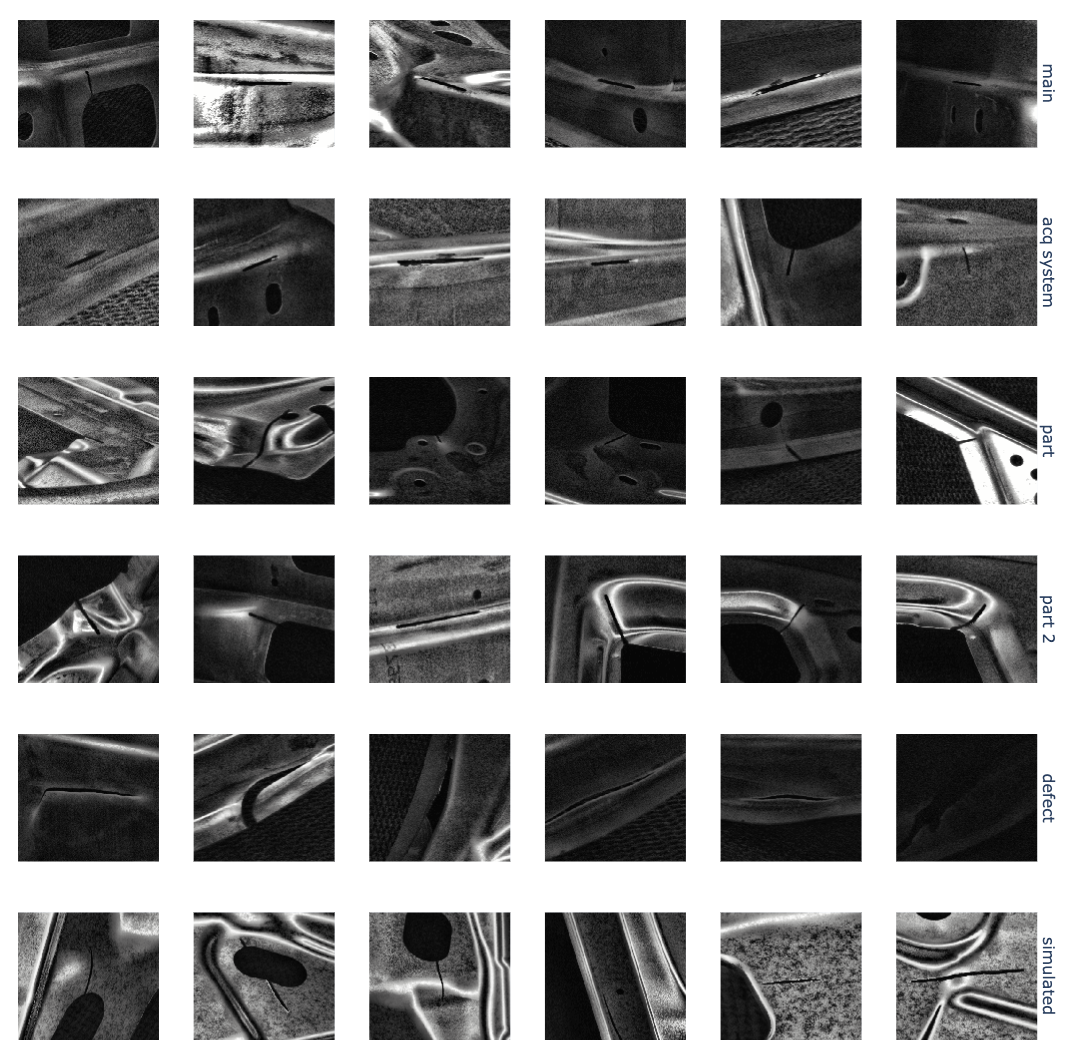}
    \caption{Comparison of some defect close-ups for the different test sets (main, change of acquisition system, two changes of part reference, change of defect aspect), as well as for the simulated data.}
    \label{fig:defect_closeups}
\end{figure}

To evaluate our models in real life conditions, we create a main real test set of 1520 images (795 defectives, 725 healthy) with different physical parts than the ones used to design the real training set. To assess generalization capabilities of our models to unseen distribution shifts, we also evaluate on four “transfer” test sets: a first one with a change of the acquisition system (new camera and lightning setup), a second one with a change of part reference (right door instead of left door), a third one with a second change of part reference (rear door instead of front door), and a last one with change in defect aspect (see Figure \ref{fig:defect_closeups}). Our four transfer test sets have respectively 1130 (285 defectives, 845 healthy), 935 (135 defectives, 800 healthy), 1245 (280 defectives, 965 healthy) and 755 (320 defectives, 435 healthy) images.

We evaluate our models on the different test sets by following the precision, recall and mAP metrics, which are all computed based on the following true positive matching criterion. Both our ground truth and our model predictions can be understood as bounding boxes. An inferred box that is intersecting an annotated ground truth box will be considered as a true positive (but only one inferred box can be matched per ground truth annotation). Inferred boxes have associated confidences, which allow to compute precision and recall at multiple thresholds.

\subsection{Training with simulated data}
\label{subsec:results}

\begin{table}[ht!]
\centering
\caption{Performance results: mAP for each trained model on all test sets after 200 epochs. Transfer test sets are \textit{acq system} (change in acquisition system), \textit{part} (change in part reference), \textit{part 2} (another change in part reference), and \textit{defect} (change in defect aspect).}
\label{tab:results}
\begin{tabular}{l||ccccc}
 &      \textit{main} &  \textit{acq system} &     \textit{part} &     \textit{part 2} &   \textit{defect} \\
               &           &             &           &           &           \\
\hline
Real                   &  0.913 &       0.452 &  0.421 &  0.432 &    0.591 \\
Simulated              &  0.316 &       0.183 &  0.402 &  0.446 &    0.099 \\
Real + additional healthy           &  0.912 &       0.500 &  0.436 &  0.444 &    0.569 \\
Simulated + additional healthy      &  0.563 &       0.337 &  0.489 &  0.447 &    0.367 \\
Real + simulated         &  0.913 &       0.664 &  0.622 &  0.560 &    \textbf{0.619} \\
Real + simulated + additional healthy &  \textbf{0.922} &       \textbf{0.699} &  \textbf{0.663} &  \textbf{0.590} &    0.610 \\
\end{tabular}
\end{table}

To answer our questions, we train models using different training data. Results after convergence are summarized in table \ref{tab:results}. Learning curves for mAP, precision, and recall on all test sets are given in Appendix \ref{appendix:details} (Figure \ref{fig:all_curves_1} and \ref{fig:all_curves_2}). We first train a baseline model using real data only. As expected, the baseline model performs very well on the main test set (mAP of 0.913). However, a significant drop in performance can be noticed when changing the content of the test set (mAP drops to approximately 0.4-0.6, depending on the type of transfer). The fact that the model trained on real data only does not generalize well to unseen distributions is mostly due to the generation of false positives, as evidenced by the precision and recall learning curves from Figure \ref{fig:all_curves_1}. Indeed, they indicate a high recall and low precision on all transfer test sets for the “Real” baseline training. In practice, the generated false positives are often due to the stamped holes in the door, which may look like cracks under certain point of views. This could explain the performance drop for the acquisition system change, but also for the two part reference changes, which both induce a lot of these stamped holes to become visible. More broadly, this lack of generalization is not too surprising, since these real images are expected to be very correlated with one another.

\subsubsection{Simulated data only}
We first compare our baseline model trained on real data with a model trained on simulated data only.
Unfortunately, using solely simulated images does not allow to get satisfatory performances on the main test set (mAP after convergence is only of 0.316). Interestingly, we nevertheless observe that it is possible to get a performance similar to the real baseline on both part change transfer test sets, using simulations only (mAP of around 0.4). This highlights the simulation data transfer capability: simulated data generalization power is approximately “one reference away”. To nuance, one can however notice that the simulation model does not transfer well to the defect aspect change, which thus appears as a harder transfer task: we hypothesize that this is due to the difference in the defect size distribution between the simulated data and the change of defect aspect test set. This answers question (a).

\subsubsection{Simulated data with additional healthy real images}
Second, we aim to assess whether we can improve the simulation data performance by using the additional real healthy images combined to the simulated defective images as the training set. Performance is improved on the main test set by more than 0.20 points (mAP of 0.563), without nonetheless getting close to the “Real + additional healthy” baseline counterpart (mAP of 0.912). In terms of generalization, performance of the simulated model improves by a significant margin on the transfer test sets when using the additional real healthy images (+0.15 points on acquisition system change, +0.09 points on the first part change, +0.27 points on the defect aspect change), even beating the “Real” previous baseline transfer capability on the first part change test set. Even though this model may still not be on par with the real baselines on the main test set, its performance on both part change test sets is actually better than the one from the same baselines. This answers question (b).

\subsubsection{Simulated data in combination with real data}
Finally, using both simulated and real defective images, performance is the same than using only real defective images on the main test set (mAP of 0.913). However, mAP is improved on all transfer test sets (from +0.03 for defect aspect change to +0.21 for the second part aspect change). We test whether this gain is simply due to the quantity of data by comparing this model to a baseline model trained on the real data plus the additional healthy real images (hence, number of images is comparable between the two trainings). The fact that the “Real + additional healthy” model does not generalize as good as the “Real + simulated” model enable us to conclude that simulations are actually helping generalization. We also observe an improvement in generalization when comparing the “Real + simulated + additional healthy” model to the “Real + additional healthy” baseline. Overall, best performances are obtained when using all available training data: real, additional healthy real, and simulated images. This answers question (c).

\subsubsection{Conclusion: simulated data helps generalizing better}
As seen in the previous sections, simulations may not improve performance on the main test set, but definitely help generalizing better on all tested unseen distributions, especially if used in combination with real images. We hypothesize this gain in generalization power to come from two phenomenons. First, a reduction of the false positives that are induced by the new domains shifts. Second, an improvement in learning the defect aspect, due to the randomized simulated defects. In the first case, we should observe a decrease in false positives (improvement of precision) when using the simulated images, while in the second case, we should observe a decrease of false negatives (improvement of recall).

\subsection{Why is generalization improved with the simulated data?}

\begin{figure}
    \centering
    \includegraphics[width=10cm]{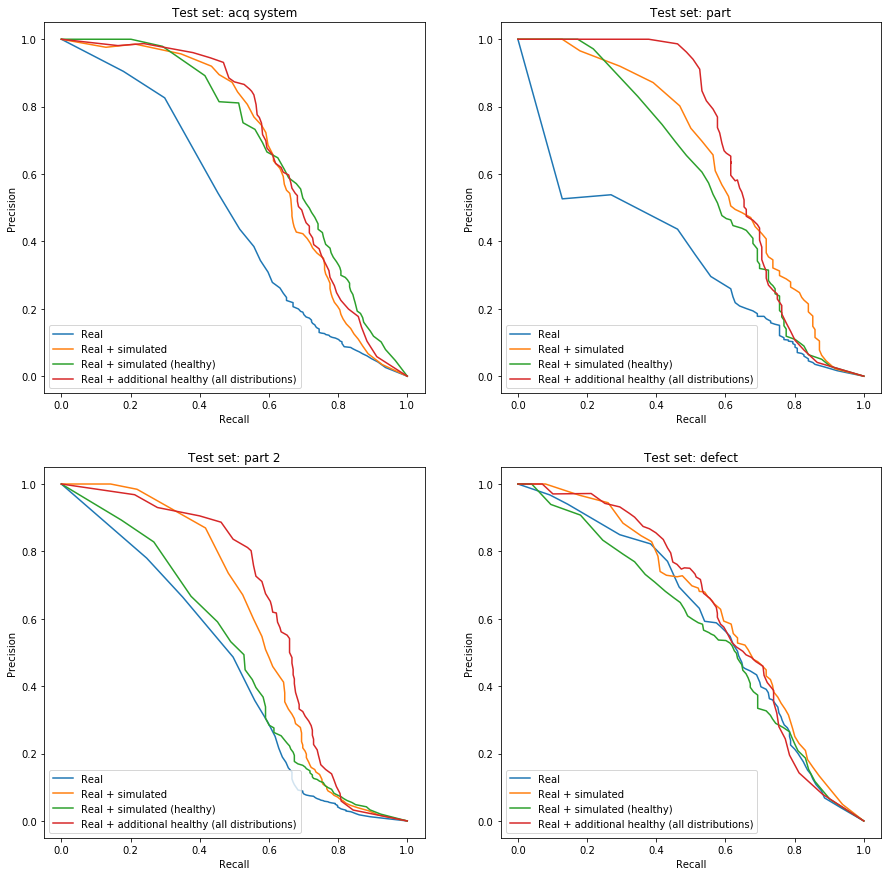}
    \caption{Precision-recall curves on all four transfer test sets for when training without simulated images (blue), with simulated images (orange), with non-defective simulated images (green), and with additional healthy real images coming from all tested distributions (red). Models are trained until convergence (200 epochs).}
    \label{fig:precision_recall}
\end{figure}

\begin{table}[ht!]
\centering
\caption{mAP comparison for when training without simulated images, with simulated images, with non-defective simulated images, and with additional healthy real images coming from all tested distributions. Models are trained until convergence (200 epochs).}
\label{tab:precision_recall}
\begin{tabular}{l||ccccc}
 &   \textit{main} & \textit{acq system} &  \textit{part} &  \textit{part 2} & \textit{defect} \\
\hline
Real                  &  0.913 &      0.452 &  0.421 &  0.432 &   0.591 \\
Real + simulated       &  0.913 &      0.664 &  0.622 &  0.560 &  0.619 \\
Real + simulated (healthy) &  \textbf{0.923} &      0.663 &  0.551 &  0.479 &   0.564 \\
Real + additional healthy (all distributions)       &  0.918 &      \textbf{0.679} &  \textbf{0.660} &  \textbf{0.627} &   \textbf{0.622} \\
\end{tabular}
\end{table}

We test the above hypotheses by comparing recall and precision on all test sets for two trainings: “Real” and “Real + simulated”. To test the hypothesis further, we train and evaluate two supplementary models. The first one is trained with the real images as well as around the same amount of healthy simulated images, and should help determine whether the simulated defects actually help improving the recall or not. The second one is trained with the real images as well as around the same amount of additional healthy real images, but this time coming from all tested distributions for which this makes sense (i.e. acquisition system change and both part changes). This second training should help assess the ability of the model trained with simulation data to improve precision. Note that such healthy images are here relatively easy to collect and annotate, given that the changes in distribution happened already in production.

Results are recapitulated in Figure \ref{fig:precision_recall} for the precision-recall curves, and in Table \ref{tab:precision_recall} for the corresponding mAPs. While precision and recall values are very similar between the “Real” (blue curves) and the “Real + simulated” (orange curves)  trainings on the main test set, both precision and recalls are improved when adding the simulated images on the transfer test sets. Replacing the defective simulated images by healthy ones has a significant negative impact on performance for most test sets, as seen in Table \ref{tab:precision_recall}. This decrease in mAP can be explained by a loss in both precision and recall, depending on the test set (see Figure \ref{fig:precision_recall}). Additionally, the model trained on the real data as well as the additional healthy data coming from all transfer distributions performs only slightly better than our “Real + simulated” model. Most of the improvement here is coming from an improvement of precision, as shown from the precision-recall curves for the two part changes test sets (see Figure \ref{fig:precision_recall}).

As a conclusion, simulated data mostly reduces the amount of false positives on unseen distributions, but also improve recall on these distributions thanks to the randomization of the simulated defects variability. Importantly, the main advantage of using simulations compared to using diverse healthy images for false positive reduction is that thanks to domain randomization, it is possible to prevent domain shift before it actually happens in production. In comparison, collecting many diverse healthy images from the production to try cover various types of shifts in advance is not expected to be an easy task, not to mention that this may be time consuming as well.


\subsection{Impact of the network size and pre-training}

Because the capacity of our custom network is limited, we decide to make it deeper in order to assess the potential role of the network size on transfer capabilities, with and without simulated images. It has indeed experimentally be shown that the network size has a role in robustness to common perturbations\cite{hendrycks2019benchmarking}, and thus in generalizing better. As a result, the number of trainable parameters for our network is roughly doubled. This bigger network is pre-trained on the ImageNet dataset classification task, and fine-tuned using either “Real + additional healthy”, or “Real + simulated + additional healthy” images for training. Results are presented in Table \ref{tab:results_deeper}. Learning curves for mAP, precision, and recall on all test sets are available in Appendix \ref{appendix:details} (Figure \ref{fig:all_curves_3}).

\begin{table}[ht!]
\centering
\caption{mAP comparison between the original and the deeper network architectures, with and without the use of simulated data. Each model is trained for 200 epochs and reaches convergence.}
\label{tab:results_deeper}
\begin{tabular}{l||ccccc}
 &   \textit{main} & \textit{acq system} &  \textit{part} &  \textit{part 2} & \textit{defect} \\
\hline
\textbf{Baseline network}\\
Real + additional healthy                       &  0.912 &      0.500 &  0.436 &  0.444 &   0.569 \\
Real + simulated + additional healthy             &  0.922 &      0.699 &  0.663 &  \textbf{0.590} &   0.610 \\
\textbf{Deeper network} \\
Real + additional healthy           &  \textbf{0.957} &      0.650 &  0.607 &  0.538 &   0.526 \\
Real + simulated + additional healthy &  0.943 &      \textbf{0.814} &  \textbf{0.717} &  0.560 &   \textbf{0.620} \\
\end{tabular}
\end{table}

When comparing the effect of a deeper network on real data alone (without simulations), one can already notice a significant gain in mAP (from +0.04 to +0.17) on all test sets, save for the change in defect aspect test set. This confirms the low capacity of our original baseline network architecture, as well as the hypothesis that bigger architectures may help for better generalization. Similarly to what was discussed in \ref{subsec:results}, although simulated images do not profit to the main test set, they still improve generalization on all transfer test sets when using the deeper network (gains varying from 0.02 to 0.16 points of mAP). One should note that the gains coming from the use of simulated data are comparable to the ones obtained via the architecture improvement alone. Interestingly, both types of gains are cumulative.

We also investigate the impact of the ImageNet pre-training on the deeper network, but find that the latter is actually null on mAP (pre-training only improves convergence -- data not shown).

\section{Conclusion}
\label{sec:conclusion}

In this work, we used a generic pipeline to massively generate synthetic training data and to assess the quality of industrial parts, via deep learning. It is first important to emphasize that iterating on the generation of simulated data is critical, but a very difficult thing to do in practice, for two reasons. First, it is almost impossible to predict \textit{a priori} how good the generated defects will be until an actual model is trained on them. Second, understanding \textit{a posteriori} why a given model fails on the test data is also in itself challenging.

We showed that while the use of simulation images alone is not sufficient (this was also the conclusion of our previous work\cite{DBLP:journals/corr/abs-2104-02980}), its combination with real images can be beneficial. This is especially true when testing the model on unseen distributions: the combined use of simulated and real images then allows better generalization almost systematically. We demonstrate that this improvement mostly comes from a reduction in false positives, even though we also observe a better learning of the real defects. Furthermore, the generalization power brought by the simulated data is approximately “one part reference away”.

As a conclusion, simulated data could therefore be used as a safeguard against changes of input image distributions, which arise regularly in production. In the future, we plan to further work on finding simpler ways to iterate on the simulated data. Simulations could also help tackling deep learning tasks for which annotation is more tedious, such as part segmentation or part pose estimation.

\bibliography{report} 
\bibliographystyle{spiebib} 

\pagebreak
\appendix
\section{Learning curves}
\label{appendix:details}

\begin{figure}[!b]
    \centering
    \includegraphics[width=15cm]{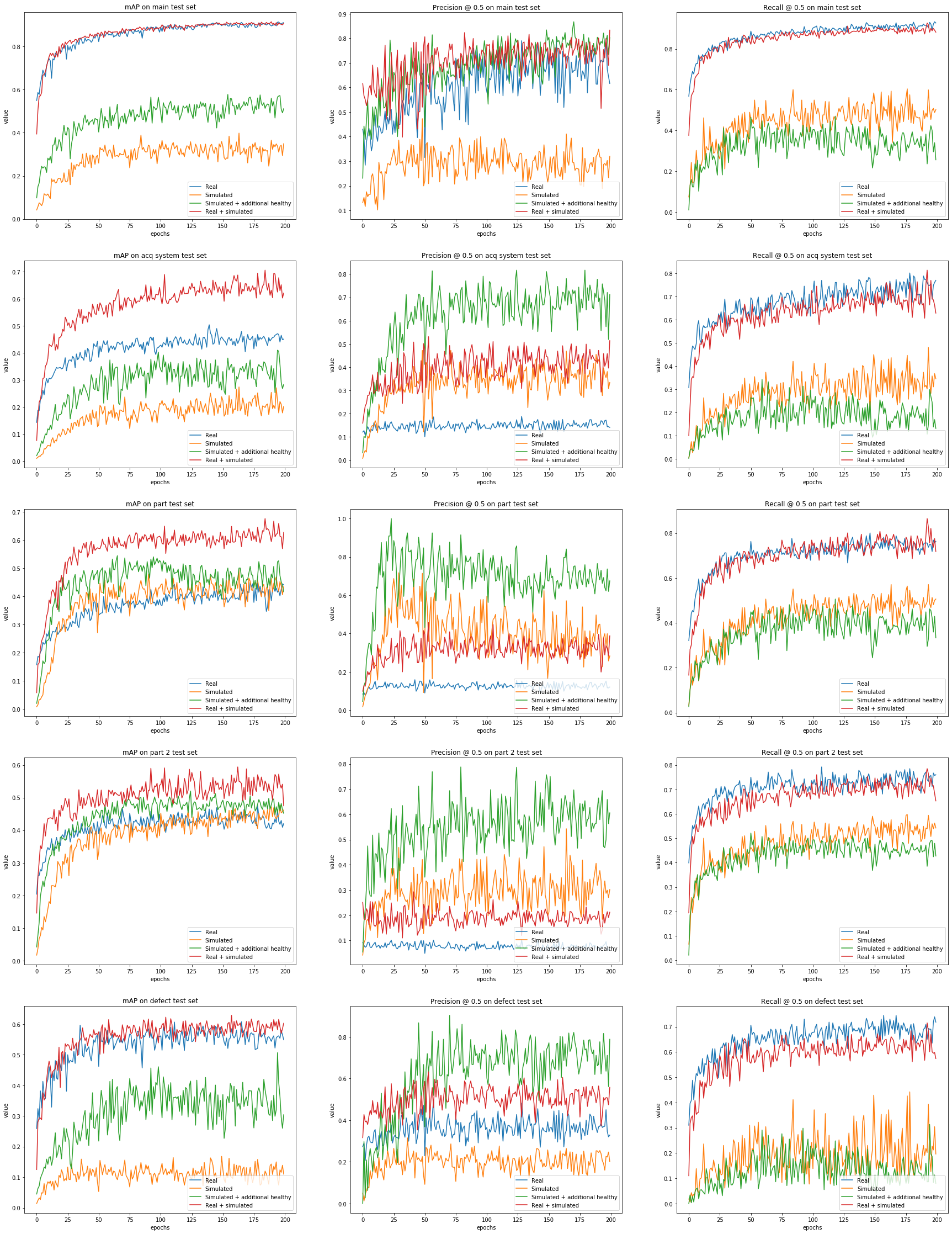}
    \caption{mAP, precision, and recall on all test sets over training epochs for trainings “Real”, “Simulated”, “Simulated + additional healthy”, and “Real + simulated”.}
    \label{fig:all_curves_1}
\end{figure}

\begin{figure}
    \centering
    \includegraphics[width=15cm]{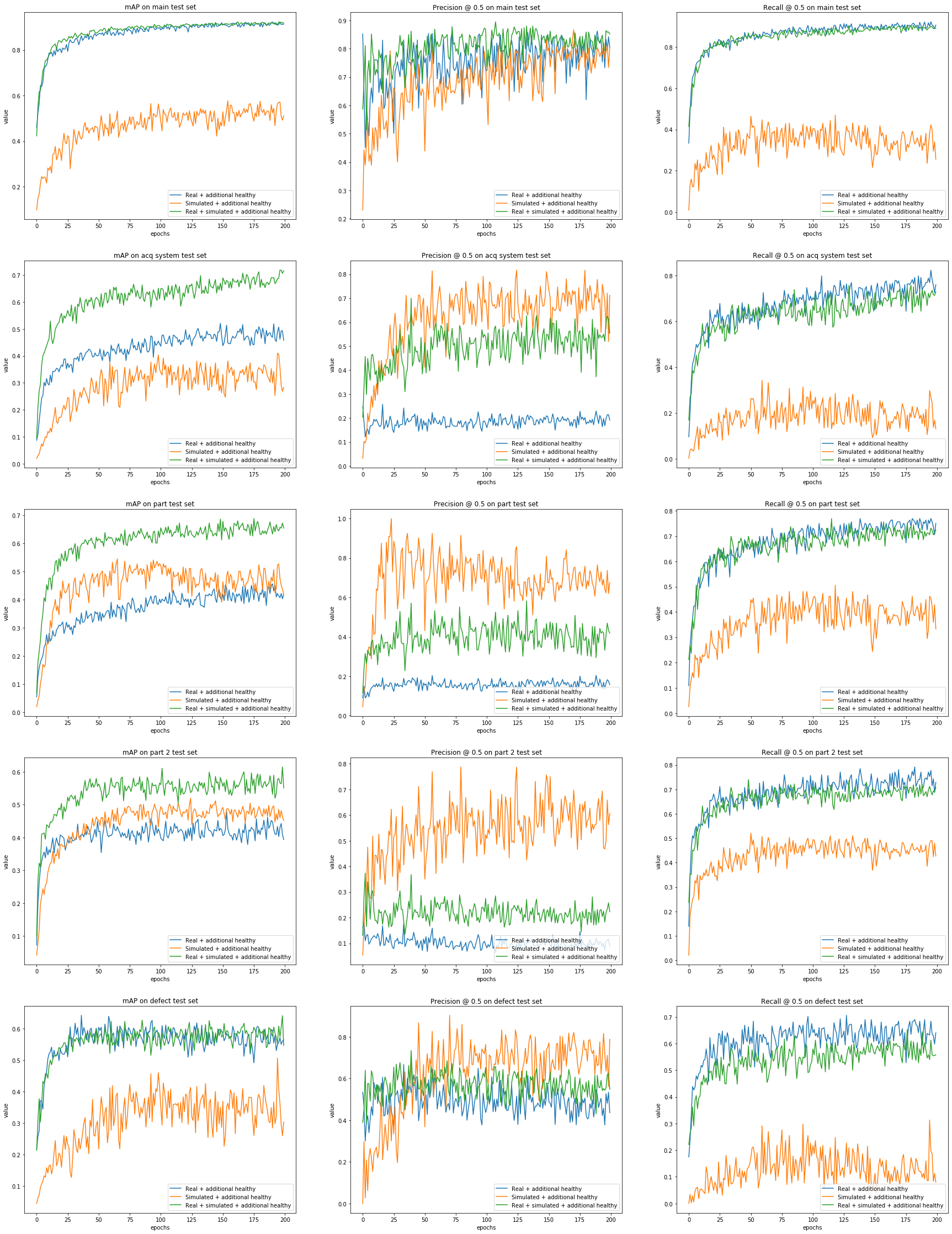}
    \caption{mAP, precision, and recall on all test sets over training epochs for trainings “Real + additional healthy”, “Simulated + additional healthy”, and “Real + simulated + additional healthy”.}
    \label{fig:all_curves_2}
\end{figure}

\begin{figure}
    \centering
    \includegraphics[width=15cm]{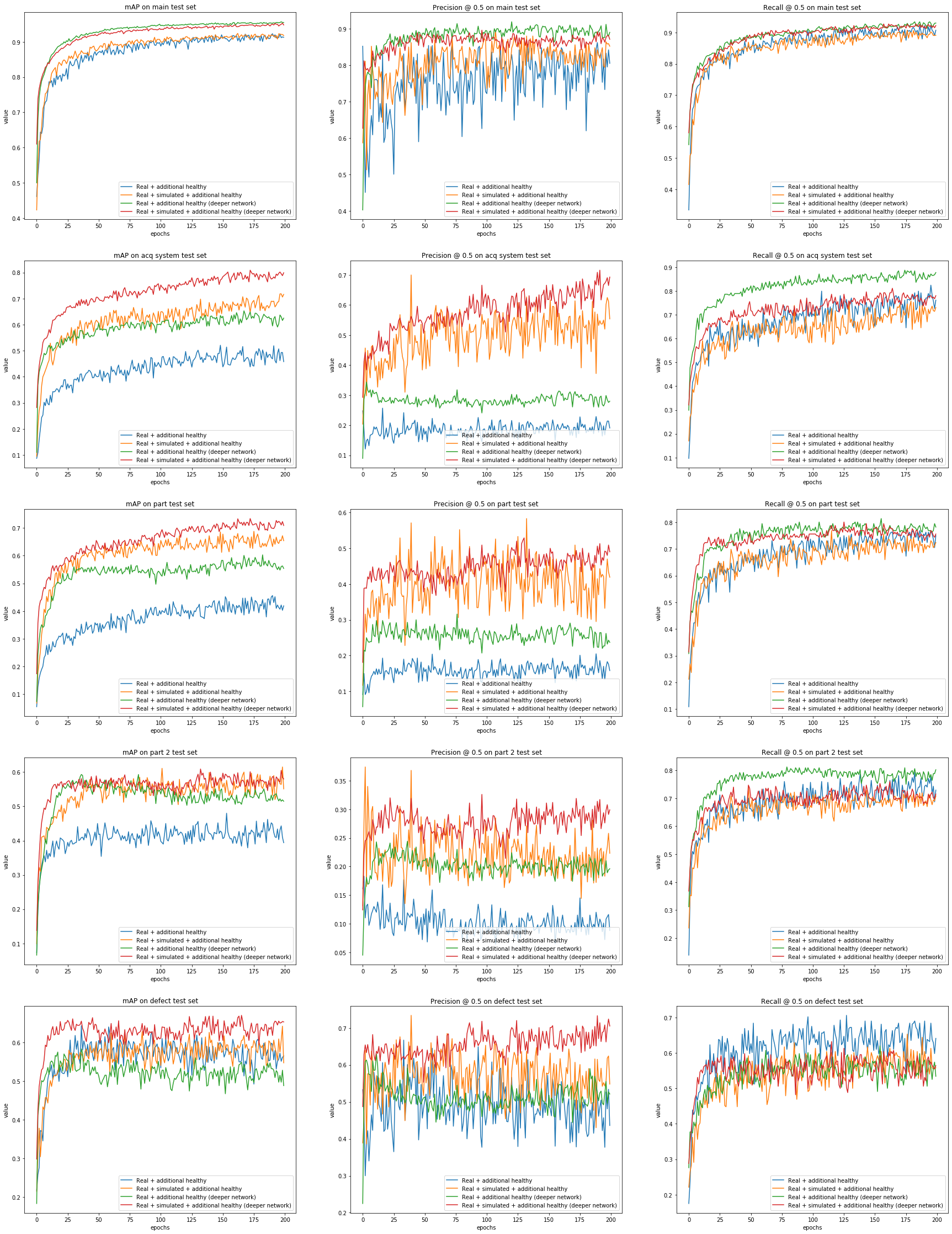}
    \caption{mAP, precision, and recall on all test sets over training epochs for trainings “Real + additional healthy” and “Real + simulated + additional healthy” for both the baseline and the deeper network architecture.}
    \label{fig:all_curves_3}
\end{figure}
\end{document}